\documentclass{INTERSPEECH2023}
\usepackage{placeins}

\interspeechcameraready

\title{System Description for the Displace Speaker Diarization Challenge 2023}
\name{Ali Aliyev$^1$}
\address{
  $^1$MTS AI, Russia
\email{a.aliyev@mts.ai}
}
\begin{document}

\maketitle
 
\begin{abstract}
This paper describes our solution for the Diarization of Speaker and Language in Conversational Environments Challenge (Displace 2023). We used a combination of VAD for finding segments with speech, Resnet architecture based CNN for feature extraction from these segments, and spectral clustering for features clustering. Even though it was not trained with using Hindi, the described algorithm achieves the following metrics: DER 27. 1\% and DER 27. 4\%, on the development and phase-1 evaluation parts of the dataset, respectively.

\end{abstract}

\section{Introduction}
Diarization is the process of separating speech belonging to different speakers. In diarization algorithms, we usually find segments with speech in the audio signals, then obtain a numerical representation (features) for each segment and then cluster the segments based on those features. We should also take into consideration that the error of each step directly affects the error of the next step.

 Diarization of Speaker and Language in Conversational Environments Challenge \cite{dicpl} addresses the problem of separating voices by speaker and by language. The peculiarity of this challenge is that, unlike other such competitions, the same speakers speak two different languages, namely English and Hindi. It is this feature that makes this challenge unique among other similar diarization challenges. In our solution, we did not use Hindi during the training, but in spite of this, we achieved good results.
This challenge consists of two tracks:
\begin{itemize}
    \item Track-1: Speaker diarization in multilingual scenarios.
    \item Track-2: Language diarization in multi-speaker settings.
\end{itemize}

In the following sections, we will describe our solution for speaker diarization track.

\section{ System description}
Usually, all algorithms for speaker diarization consist of three parts:
\begin{enumerate}
\item Voice activity detector
\item Feature extractor
\item Clustering algorithm
\end{enumerate}

\subsection{Voice activity detector}

VAD is present in almost all diarization algorithms \cite{park22e_interspeech,dissen22_interspeech}, because segments with noise or other extraneous sounds can lead to an error in the following steps. The clustering algorithm, which independently selects the number of clusters (speakers) in the audio files, may make a false positive prediction and create a cluster for an extra speaker that is not actually there.
 
We selected the pre-trained Silero VAD v4\footnote{\url{https://github.com/snakers4/silero-vad}} model, which is one of the most accurate open-source solution for the speech activity detection task. This model achieves a ROC-AUC score equal to 0.9 on the Libryparty dataset \cite{speechbrain} and 0.99 on the AVA speech activity dataset \cite{chaudhuri18_interspeech}. Although the v4 version of the model achieves better scores on these datasets, while v3 achieves 0.87 and 0.93 respectively, the switch from v3 to v4 did not significantly affect our metrics.

However, after experiments, we were not completely satisfied with the results of Silero VAD on this task, so we decided to test another solution as well. We took WebRTC VAD\footnote{\url{https://github.com/wiseman/py-webrtcvad}}, which achieves 0.81 on the Libryparty dataset and 0.66 on the AVA speech activity dataset.

\subsection{Feature extractor for speaker recognition}
Our feature extractor was originally trained for speaker verification, but it is just as well suited for the speaker diarization task.
\subsubsection{Training data}

Like most pipelines for speaker verification, our neural network uses the VoxCeleb2 \cite{chung18b_interspeech} dataset, which contains 1,092,009 utterances and 5,994 speakers, as its main training dataset. But, since the main goal of this competition is to apply speaker diarization systems in a multilingual environment, the feature extractor should also be trained in two languages.
However, we did not have enough data in Hindi. To solve this problem, we took advantage of CNN's training feature for speaker verification, where training in two languages, improves metrics in most other languages. This ability to adapt for other languages is also supported in other works \cite{10.1007/978-3-031-06791-4_20}. 
So we took Common Voice Corpus 12.0 \cite{commonvoice:2020} for Russian and combined this dataset with VoxCeleb2.

In the end, we got a dataset with 2600 hours of speech in English and 229 hours of speech in Russian. 
\subsubsection{Model architecture and training}

We used Resnet \cite{7780459} models as the basic architecture. Namely, Resnet-34 and Resnet-293. And as input data, we used fixed \SI{2}{\second} segments, which were randomly cut from utterances from our dataset. From which we further extract 80-dimensional MEL f-banks with a window length of \SI{25}{\milli\second} and \SI{10}{\milli\second} stride length. For data augmentation, we used the Music, Speech, and Noise Corpus (MUSAN) \cite{musica} to add noise, music, and other extraneous sounds and reverberation from the Room Impulse Response and Noise Database (RIR) \cite{ko2017study}. 
AAM-Softmax Loss \cite{8953658} was used to train the model. It took us 18 hours to train Resnet-34 and 97 hours to train Resnet-293 on 8 NVIDIA Tesla A100 40 GB GPUs. Each model has been trained during 150 epochs.




\subsection{Overlapped speech detection}

One of the significant problems of all speaker diarization systems is speech segments where there are voices of two or more speakers. In such segments, the feature extractor produces incorrect embeddings due to the presence of two or more voices. Usually additional classifiers are used to detect such segments, but we went the other way. Instead of using additional detection methods \cite{9053096,4518619}, we divided all the segments after VAD into additional subsegments, where from each of them we extract the features and only then we clustered them. To get subsegments, we will use the sliding window technique with a window length of \SI{2}{\second} and \SI{0.4}{\second} stride length, it is with these parameters that we achieved the best results.

\subsection{Clustering}

We used Spectral Clustering \cite{ng2001spectral} as a clustering algorithm. Because it can work under more difficult conditions than other clustering methods, such as k-means etc. Here's how it works: \\

\begin{enumerate}
\item Calculating of a similarity matrix for our embeddings from the feature extractor. We use cosine similarity as a similarity measure.



%
\item Calculating of a Laplacian matrix from a similarity matrix.
\item Then we're solving a standard eigenvalue problem for a real symmetric matrix to calculate eigenvectors and eigenvalues of Laplacian matrix.

\item To solve the problem of determining the $k$ (number of clusters), we used a heuristic method \cite[pp.\ 410--411]{scl} based on eigenvalues.

\item After we have computed $k$, we can now apply k-means clustering to the first $k$ eigenvectors from the previous steps.
\end{enumerate}

\section{Challenge dataset}

The dataset for this contest consists of 3 parts:
\begin{itemize}

\item Development dataset with ground truth labels. This part contains 27 audio files in \emph{wav} format and annotations in \emph{rttm} format. The total duration of the utterances is 15 hours and 45 minutes, and  the most of the files are 30 minutes long, and some are about an hour long. Usually all files are single-channel, but one file, namely M043.wav, was in stereo for some reason. So we fixed that by converting it to a mono channel audio file. The maximum number of speakers found in the ground truth files was 4.

\item Phase 1 evaluation dataset contains 20 audio files in wav format, but without annotation. The total duration of the files is 11 hours and 24 minutes. Most of the files are also 30 minutes long, and some are about an hour long. As in the previous part of the dataset, this one also has one file (M053.wav) recorded in stereo channel. 

\item At the time of writing, this article phase 2 evaluation dataset was not yet available for participants.
\end{itemize}

Since this contest contains two tracks, each track has its own annotations, but both tracks are using the same audio files. 
\section{Experiments results}

In this section, we will present the results of our tests with different  parts of our algorithm.

As a metric to calculate error, organizers use a metric called diarization error rate (DER).  This error rate is the sum of the following values:
\begin{itemize}

\item Speaker error (SE) - percentage of scored time for which the wrong speaker ID is assigned for a speech segment.

\item False alarm speech (FA) - percentage of scored time where non-speech segment was incorrectly marked as a segment which contains speech.

\item Missed speech (MS) - percentage of scored time where a segment with speech was incorrectly marked as non-speech segment.
\end{itemize}

Based on the above, we can draw the following conclusions. The speaker error is directly affected by the feature extractor and the clustering algorithm. And false alarm speech and missed speech depend on the quality of the VAD.
The closer the error is to 0, the better for us. DER may also exceed 100, since it is the sum of several errors.

Note that the authors of the competition use an implementation called dscore\footnote{\url{https://github.com/nryant/dscore}} to calculate DER metric, so depending on the version of the implementation used to calculate the metrics, the numbers may be slightly different.

\subsection{Voice activity detector}

The tables below will show all three types of error, on the basis of which the DER is formed. However, primarily, to evaluate the quality of VAD performance, we need to look at false alarm speech (FA) and missed speech (MS). All VAD experiments were performed under identical conditions, with a window length of \SI{2}{\second} and \SI{0.4}{\second} stride length for the feature extractor (Resnet-34) and with spectral clustering. 

\subsubsection{Silero VAD}

As we can see in Table~\ref{tab:vad1}, Silero VAD 4.0 performs a slightly better than the previous version on this competition dataset. But according to the available information from the authors of Silero VAD, the fourth version of the model is 3.4\% better in the AVA Spoken Activity Dataset and 6.1\% better in the Libryparty dataset. However, the gain on the displace2023\_dev dataset was only 0.6\%.

\begin{table}[h]
  \caption{The results of different versions of Silero VAD, with the same parameters on the displace2023\_dev dataset.}
  \label{tab:vad1}
  \centering
    \begin{tabular}{ l c c c c c c c c c }
    \toprule
    \multicolumn{1}{c}{\textbf{VAD version}} &
    
    \multicolumn{1}{c}{\textbf{MS}} &
    \multicolumn{1}{c}{\textbf{FA}} &
    \multicolumn{1}{c}{\textbf{SE}} &
    \multicolumn{1}{c}{\textbf{DER}} \\

    \midrule
    Silero-VAD 3.1 & $17.4$ & $5.4$ & $4.1$ & $26.9$ \\
    Silero-VAD 4.0 & $17.3$ & $5.4$ & $4.1$ & $26.8$  \\
    \bottomrule
  \end{tabular}
\end{table}

In Table~\ref{tab:vad2} we can see that we had to lower the threshold a lot in order to improve our metrics. This is due to the fact that Silero VAD is trained on a multilingual dataset that does not include Hindi, and since in addition to English the speakers in this dataset also speak Hindi, this worsens the accuracy of VAD. And because we lowered the threshold, it led to a lot of false-positive and false-negative segments. And in some examples there were so many of them, that the algorithm simply returns only one segment with timestamps of the beginning and end of the audio file. Apparently, there is some kind of algorithm that combines the overlapping segments.

\begin{table}[h!]
  \caption{Silero VAD 4.0 results with different thresholds on the displace2023\_dev dataset.}
  \label{tab:vad2}
  \centering
    \begin{tabular}{ c c c c c c c c c c }
    \toprule
    \multicolumn{1}{c}{\textbf{Threshold}} &
    
    \multicolumn{1}{c}{\textbf{MS}} &
    \multicolumn{1}{c}{\textbf{FA}} &
    \multicolumn{1}{c}{\textbf{SE}} &
    \multicolumn{1}{c}{\textbf{DER}} \\

    \midrule
    0.15 & $17.3$ & $5.4$ & $4.1$ & $26.8$ \\
    0.25 & $25.4$ & $2.8$ & $3.0$ & $31.2$  \\
    0.50 & $30.5$ & $2.1$ & $2.6$ & $35.2$  \\
    0.75 & $35.4$ & $1.7$ & $2.4$ & $39.5$  \\

    \bottomrule
  \end{tabular}
\end{table}
\FloatBarrier
\subsubsection{WebRTC VAD}
WebRTC VAD uses a concept such as aggressiveness instead of threshold. This parameter affects the sensitivity level of non-speech segments filtering. There are 4 levels of aggressiveness from 0 to 3, where 0 is the least sensitive and 3 is the most sensitive.
Table~\ref{tab:vad3} shows that the more aggressive the WebRTC VAD is, the more speech we start to skip in the dataset. 

It accepts audio segments with durations: 10, 20 and \SI{30}{\milli\second}. As we see from Table~\ref{tab:vad4}, the best result is obtained if we divide the original audio signal into \SI{20} {\milli\second} segments. The tests were made using zero level aggressiveness.

\begin{table}[h!]
  \caption{WebRTC VAD results with different level of aggressiveness on the displace2023\_dev dataset.}
  \label{tab:vad3}
  \centering
    \begin{tabular}{ c c c c c c c c c c }
    \toprule
    \multicolumn{1}{c}{\textbf{Aggressiveness}} &
    
    \multicolumn{1}{c}{\textbf{MS}} &
    \multicolumn{1}{c}{\textbf{FA}} &
    \multicolumn{1}{c}{\textbf{SE}} &
    \multicolumn{1}{c}{\textbf{DER}} \\

    \midrule
    0 & $19.2$ & $4.5$ & $3.7$ & $27.4$ \\
    1 & $20.0$ & $4.3$ & $3.6$ & $27.9$  \\
    2 & $21.6$ & $4.0$ & $3.4$ & $29.0$  \\
    3 & $28.7$ & $2.8$ & $3.1$ & $34.6$  \\

    \bottomrule
  \end{tabular}
\end{table}

\begin{table}[h!]
  \caption{WebRTC VAD results with different lengths of input segments on the displace2023\_dev dataset.}
  \label{tab:vad4}
  \centering
    \begin{tabular}{ c c c c c c c c c c }
    \toprule
    \multicolumn{1}{c}{\textbf{Duration}} &
    \multicolumn{1}{c}{\textbf{MS}} &
    \multicolumn{1}{c}{\textbf{FA}} &
    \multicolumn{1}{c}{\textbf{SE}} &
    \multicolumn{1}{c}{\textbf{DER}} \\

    \midrule
    10 & $19.3$ & $4.5$ & $3.7$ & $27.5$ \\
    20 & $19.1$ & $4.5$ & $3.7$ & $27.3$ \\
    30 & $19.2$ & $4.5$ & $3.7$ & $27.4$ \\
    \bottomrule
  \end{tabular}
\end{table}

As we see with WebRTC VAD DER is equals to 27.3\%, and with Silero VAD DER is 26.8\% on the development part of the dataset. We could say that the correlation of results will be the same if we compare these two algorithms with each other on the evaluation part of our dataset. But it turned out to be the opposite. On the evaluation part of our dataset, our final algorithm with Silero VAD, DER was 28.2\%, and with WebRTC VAD DER dropped to 27.4\%. These results were achieved using Resnet-293 and spectral clustering. Unfortunately, more detailed metrics are not available, since we do not have ground truth files for the evaluation part of the dataset and the first phase of the competition was already closed.

\FloatBarrier
\subsection{Feature extractor}
 We used Silero VAD and spectral clustering in the all following experiments. From the Table~\ref{tab:resnet1} we see that bilingual learning does increase the accuracy of our neural network in other languages. Even though we trained on a combination of English and Russian dataset, it still showed some gain. The maximum gain would most likely be from training by using English and Hindi.

\begin{table}[h]
  \caption{Comparison of Resnet-34 trained on Voxceleb-2 alone and on Voxceleb-2+Common Voice Russian corpus on the displace2023\_dev dataset.}
  \label{tab:resnet1}
  \centering
    \begin{tabular}{ l c c c c c c c c c } 
    \toprule
    \multicolumn{1}{c}{\textbf{Dataset}} &
    
    \multicolumn{1}{c}{\textbf{MS}} &
    \multicolumn{1}{c}{\textbf{FA}} &
    \multicolumn{1}{c}{\textbf{SE}} &
    \multicolumn{1}{c}{\textbf{DER}} \\

    \midrule
    Voxceleb-2 & $17.3$ & $5.4$ & $4.2$ & $26.9$ \\
    Combined dataset & $17.3$ & $5.4$ & $4.1$ & $26.8$ \\
    \bottomrule
  \end{tabular}
\end{table}

As you know, one of the main datasets for diarization is the voxconverse \cite{Chung20} dataset, and on this dataset we achieved the best results (DER 7.2 \%) using a sliding window length of \SI{1.5}{\second} with \SI{0.75}{\second} stride length. But in this task, the best results were \SI{2}{\second} segments with \SI{0.4}{\second} steps. We can observe this on Table~\ref{tab:resnet2}. Perhaps the choice of window length depends on the linguistic features of each language and the speed of pronunciation of words.

\begin{table}[h!]
  \caption{Comparison of sliding window parameters for feeding data to the Resnet-34 input on the displace2023\_dev dataset.}
  \label{tab:resnet2}
  \centering
    \begin{tabular}{ l c c c c c c c c c }
    \toprule
    \multicolumn{1}{c}{\textbf{Parameters}} &
    
    \multicolumn{1}{c}{\textbf{MS}} &
    \multicolumn{1}{c}{\textbf{FA}} &
    \multicolumn{1}{c}{\textbf{SE}} &
    \multicolumn{1}{c}{\textbf{DER}} \\

    \midrule
    size=1.5, step=0.40 & $17.3$ & $5.4$ & $4.8$ & $27.5$  \\
    size=1.5, step=0.50 & $17.3$ & $5.4$ & $4.9$ & $27.6$  \\
    size=1.5, step=0.75 & $17.3$ & $5.4$ & $5.1$ & $27.8$  \\
    
    size=2.0, step=0.40 & $17.3$ & $5.4$ & $4.1$ & $26.8$  \\
    size=2.0, step=0.50 & $17.3$ & $5.4$ & $4.2$ & $26.9$  \\
    size=2.0, step=0.75 & $17.3$ & $5.4$ & $4.3$ & $27.0$  \\

    \bottomrule
  \end{tabular}

\end{table}

Initially we used Resnet-34 in our solution, but then we decided to increase the size of our model to Resnet-293 in order to maximize results from the feature extractor part of our algorithm. However, as you can see in Table~\ref{tab:resnet3}, we could not achieve serious improvements in the metrics. This is most likely due to the fact that the accuracy of our VAD to solve the problem for this dataset is insufficient. The VAD we chose was trained on other languages, so when we used it with Hindi, the results were worse than with the languages used during the training. And since we extract features from segments passed after the VAD, this also affects the results of the feature extractor. Changing to a better VAD, could also improve the results of the feature extractor.

\begin{table}[h!]
  \caption{Comparison of Resnet-34 and Resnet-293 trained on Voxceleb-2+Common Voice Russian corpus on the displace2023\_dev dataset.}
  \label{tab:resnet3}
  \centering
    \begin{tabular}{ l c c c c c c c c c }
    \toprule
    \multicolumn{1}{c}{\textbf{Model}} &
    
    \multicolumn{1}{c}{\textbf{MS}} &
    \multicolumn{1}{c}{\textbf{FA}} &
    \multicolumn{1}{c}{\textbf{SE}} &
    \multicolumn{1}{c}{\textbf{DER}} \\

    \midrule
    Resnet-34 & $17.3$ & $5.4$ & $4.1$ & $26.8$ \\
    Resnet-293 & $17.3$ & $5.4$ & $3.8$ & $26.5$ \\
    \bottomrule
  \end{tabular}
\end{table}

\FloatBarrier
\subsection{Clustering}

As we can see in Table~\ref{tab:sc1}, spectral clustering works better than agglomerative hierarchical clustering (AHC) \cite{DBLP:series/utcs/Nielsen16}. For AHC, we used cosine similarity with a threshold value equal to 0.5 to calculate the distance between samples, group average linkage as a linkage function, and silhouette score to determine the optimal number of clusters.

\begin{table}[h!]
  \caption{Comparison of spectral clustering with agglomerative hierarchical clustering on the displace2023\_dev dataset. Resnet-34 and Silero VAD were used.}
  \label{tab:sc1}
  \centering
    \begin{tabular}{ l c c c c c c c c c }
    \toprule
    \multicolumn{1}{c}{\textbf{}} &
    
    \multicolumn{1}{c}{\textbf{MS}} &
    \multicolumn{1}{c}{\textbf{FA}} &
    \multicolumn{1}{c}{\textbf{SE}} &
    \multicolumn{1}{c}{\textbf{DER}} \\

    \midrule
    SC & $17.3$ & $5.4$ & $4.1$ & $26.8$ \\
    AHC & $17.3$ & $5.4$ & $16.2$ & $38.9$ \\
    \bottomrule
  \end{tabular}
\end{table}

\FloatBarrier
\subsection{Final submission}
In our final submission, we decided to use a combination of Web-RTC VAD with aggressiveness equal to 0 and with segments length \SI{20} {\milli\second}, Resnet-293 with a sliding window length of \SI{2}{\second} and \SI{0.4}{\second} stride length, and spectral clustering. Table~\ref{tab:final} below shows the results on the displace 2023 development and phase-1 evaluation dataset.

\begin{table}[h!]
  \caption{Final submission results for development and evaluation phase 1 datasets}
  \label{tab:final}
  \centering
    \begin{tabular}{ l c c c c c c c c c }
    \toprule
    \multicolumn{1}{c}{\textbf{Dataset}} &
    
    \multicolumn{1}{c}{\textbf{MS}} &
    \multicolumn{1}{c}{\textbf{FA}} &
    \multicolumn{1}{c}{\textbf{SE}} &
    \multicolumn{1}{c}{\textbf{DER}} \\

    \midrule
    displace2023\_dev & $19.1$ & $4.5$ & $3.5$ & $27.1$ \\
    displace2023\_eval\_1 & X & X & X & $27.4$  \\
    \bottomrule
  \end{tabular}
\end{table}

\FloatBarrier
\section{Conclusions}
In this paper, we have described our approach for speaker diarization. Our proposed method using a combination of Web-RTC VAD, Resnet-293 and spectral clustering achieves good results, but the VAD part of the algorithm needs further improvements. Our final submission DER was 27.1\% and 27.4\%, on development and phase-1 evaluation parts of the dataset, respectively.

\bibliographystyle{IEEEtran}
\bibliography{mybib}

\end{document}